\documentclass[journal]{IEEEtran}

\usepackage{tikz}
\usepackage{caption}  
\usepackage{graphicx}
\usepackage{booktabs}
\usepackage{subcaption}
\usepackage{soul}
\usepackage{xcolor}
\usepackage[numbers]{natbib}

\usepackage[accsupp]{axessibility}  
\usepackage{todonotes}
\usepackage{tikz}
\usepackage{multirow}

\newcommand{\KITTIo}{KITTI\textsubscript{o}}

\newcommand{\KITTIt}{KITTI\textsubscript{t}}

\newcommand\vect[1]{\ensuremath{\mathbf{#1}}}

\usepackage{pifont}
\newcommand{\cmark}{\ding{51}}%
\newcommand{\xmark}{\ding{55}}
\usepackage{tabularx}
\usepackage{makecell}
\usepackage{colortbl}
\usepackage{stmaryrd}
\usepackage{stmaryrd}
\usepackage{algorithm}
\usepackage{algpseudocode}

\usepackage[pagebackref,breaklinks,colorlinks,citecolor=blue]{hyperref}

\usepackage{orcidlink}

\usepackage{balance}

\begin{document}

\title{Let It Flow: Simultaneous Optimization of 3D Flow and Object Clustering}

\author{Patrik Vacek$^{1}$,
\IEEEmembership{Member, IEEE},
\and
David Hurych$^{2}$,
\IEEEmembership{Member, IEEE},
\and
Karel Zimmermann$^{1}$
\IEEEmembership{Member, IEEE},
\and
Tomáš Svoboda$^{1}$, 
\IEEEmembership{Member, IEEE}
\thanks{The research leading to these results has received funding from the Czech Science Foundation under Project GA 24-12360S.
This work was co-funded by the European Union under the project
Robotics and advanced industrial production (reg. no. CZ.02.01.01/00/22\_008/0004590).
This research received the support of EXA4MIND project, funded by a European Union´s Horizon Europe Research and Innovation Programme, under Grant Agreement N° 101092944. Views and opinions expressed are however those of the author(s) only and do not necessarily reflect those of the European Union or the European Commission. Neither the European Union nor the granting authority can be held responsible for them
P. Vacek was also supported by 
Grant Agency of the CTU in Prague under Project SGS24/096/OHK3/2T/13. }
\thanks{$^{1}$The authors are with the Department of Cybernetics, Faculty of Electrical Engineering, Czech Technical University in Prague, 166 36 Prague, Czech Republic (e-mail: vacekpa2@fel.cvut.cz; zimmerk@fel.cvut.cz; svobodat@fel.cvut.cz), \emph{(Corresponding author: P.~Vacek)}}
\thanks{$^{2}$ David Hurych is with Valeo.ai, david.hurych@valeo.com}
}


%
\markboth{Journal of IEEE TRANSACTIONS ON INTELLIGENT VEHICLES~2024}%
{How to Use the IEEEtran \LaTeX \ Templates}

\maketitle

\begin{abstract}
We study the problem of self-supervised 3D scene flow estimation from real large-scale raw point cloud sequences. The problem is crucial to various automotive tasks like trajectory prediction, object detection, and scene reconstruction. In the absence of ground truth scene flow labels, contemporary approaches concentrate on deducing and optimizing flow across sequential pairs of point clouds by incorporating structure-based regularization on flow and object rigidity. The rigid objects are estimated by a variety of 3D spatial clustering methods. While state-of-the-art methods successfully capture overall scene motion using the Neural Prior structure, they encounter challenges in discerning multi-object motions. We identified the structural constraints and the use of large and strict rigid clusters as the main pitfall of the current approaches, and we propose a novel clustering approach that allows for a combination of overlapping soft clusters and non-overlapping rigid clusters. Flow is then jointly estimated with progressively growing non-overlapping rigid clusters together with fixed-size overlapping soft clusters. We evaluate our method on multiple datasets with LiDAR point clouds, demonstrating superior performance over the self-supervised baselines and reaching new state-of-the-art results. Our method excels in resolving flow in complicated dynamic scenes with multiple independently moving objects close to each other, including pedestrians, cyclists, and other vulnerable road users. Our codes are  publicly available on \url{https://github.com/ctu-vras/let-it-flow}. 
\end{abstract}

\begin{figure}[!t]
\includegraphics[width=0.99\linewidth]{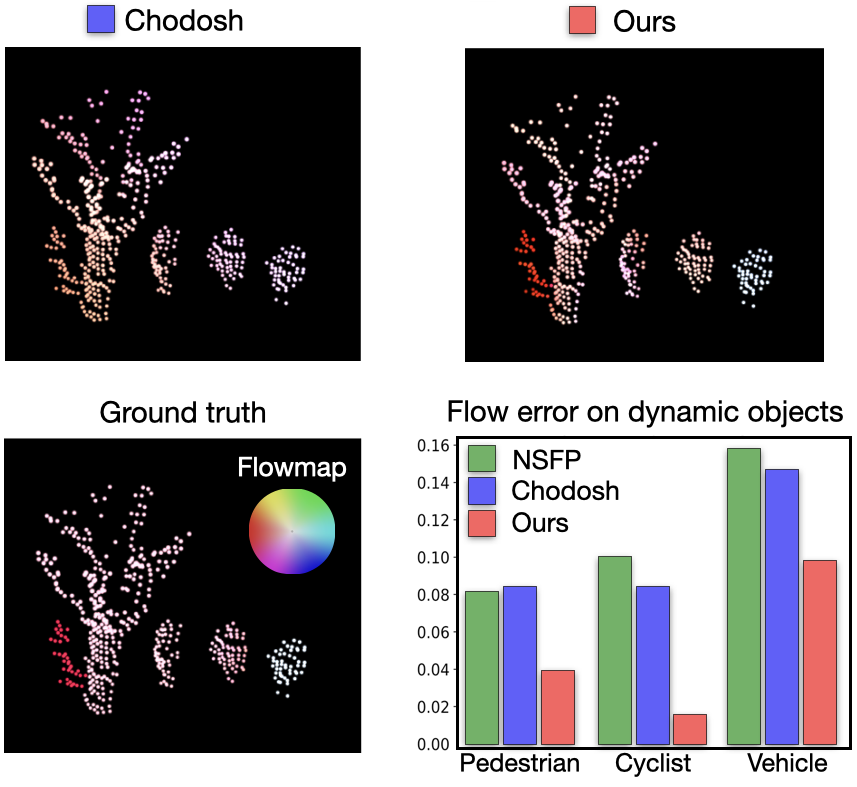}
\caption{Performance comparison of the proposed method with self-supervised competitors on Argoverse2 Dataset~\cite{Argoverse}. Our method is able to distinguish the different motion patterns and separate objects, while other methods~\cite{Chodosh2023reevaluating, vidanapathirana2023mbnsf,Li2023Fast,li2021neural} tend to under-segment objects and fit incorrect rigid motion. The qualitative example comes from Waymo~\cite{ScalableWaymo2022} dataset.
}
\label{fig:intro}
\end{figure}

\section{Introduction}
\label{sec:intro}



\IEEEPARstart{A}{ccurate} estimation of the 3D flow field~\cite{FlowNet,puy20flot,li2022rigidflow,HPLFlowNet} between two consecutive point cloud scans is crucial for many applications in autonomous driving and robotics as it provides per-point motion features inherently useful for a myriad of subsequent tasks, including semantic/instance segmentation~\cite{Baur2021ICCV,song2022ogc,3D-MOD}, object detection~\cite{detflowpretrain2022,3D-MOD}, motion prediction~\cite{lanegcnUrtasun2020,chen2022ral,ScalableWaymo2022} or scene reconstruction~\cite{Li2023Fast}. Since manually annotated data for fully-supervised training are expensive and largely not available, we focus on the self-supervised (optimization-based) setup, which is practically useful and reaches competitive results.

Self-supervision in 3D flow estimation always stems from strong prior assumptions about the scene structure. Many existing approaches, as exemplified by works~\cite{Baur2021ICCV,gojcic2021weakly3dsf,3D-MOD,li2022rigidflow,song2022ogc}, assume that the underlying scene consists of a static background and several rigid spatially compact objects, the motion of which satisfies kinematic and dynamic constraints.
State-of-the-art methods~\cite{gojcic2021weakly3dsf,li2022rigidflow,song2022ogc,Chodosh2023reevaluating,vidanapathirana2023mbnsf} typically construct spatially compact clusters~\cite{dbscan}, 
that are assumed to correspond to rigidly moving objects as a first step, and then they apply rigid flow regularization to these clusters.
Such premature clustering inevitably suffers from \emph{under-segmentation} (i.e., merging several rigid objects into a single cluster) and over-segmentation (i.e., disintegration of a single rigid object into multiple clusters).
To partially suppress the under-segmentation issue, the outlier rejection technique has recently been proposed~\cite{vidanapathirana2023mbnsf}.
Nevertheless, the quality of the initial clustering remains the main bottleneck, which prevents estimating the accurate flow, especially in situations where multiple small objects move independently close to each other. 
In contrast to existing approaches, we generate many small \emph{overlapping spatio-temporal rigid-cluster hypotheses} and then jointly optimize the flow with the rigid-body segmentation. We quantitatively and qualitatively demonstrate that such an approach mitigates the over and under-segmentation issues and consequently yields superior results, especially on dynamic objects; see Figure~\ref{fig:intro} for the comparison.
\begin{figure*}[t]
    \centering
    \includegraphics[width=0.99\linewidth]{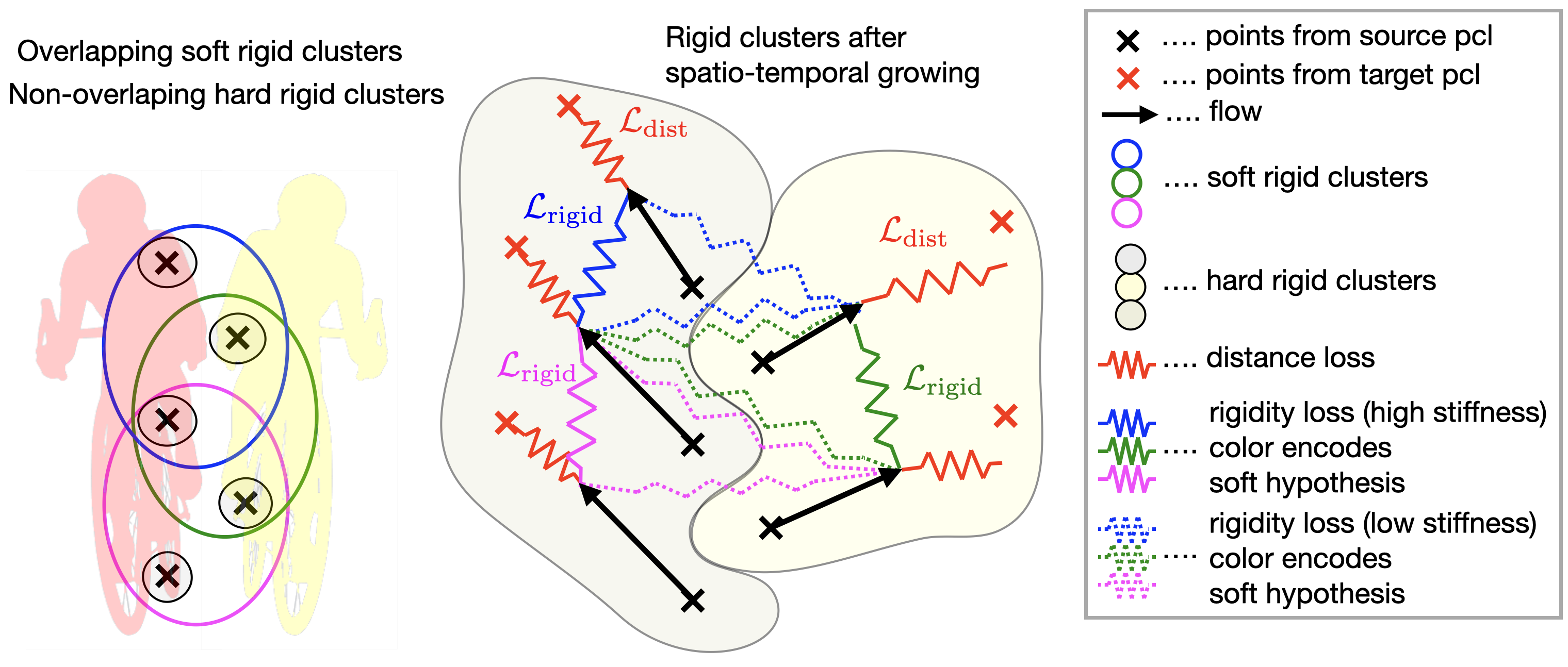}
    \caption{\textbf{Outline of the proposed losses:} 
    The left image shows a point cloud, represented by black crosses, containing two vertical objects; the silhouettes were added for visualization. No spatial clustering would segment the objects correctly; therefore, any independent flow estimation will be strongly biased by the incorrect clustering. In contrast, we cover the point cloud by (i) non-overlapping hard rigid clusters and (ii) overlapping soft rigid clusters. The right image demonstrates losses, visualized by springs, used for the flow estimation. The resulting flow is used to merge the hard clusters in the spatio-temporal domain. The procedure is repeated until convergence. The resulting hard rigid clusters deliver rigid object segmentation.
    } 
    \label{fig:losses}
\end{figure*}
%

In the proposed optimization-based framework, we employ two losses that are visualized using the analogy of a mechanical machine\footnote{Since both losses minimize the L2-norm of some quantities, we can visualize their influence on the estimated flow by ideal springs. In the ideal spring, the total conserved energy is proportional to the square of its deformation; therefore, the value of L2-loss corresponds to its energy.} with springs, see Figure~\ref{fig:losses}. The equilibrium of this machine corresponds to the globally optimal flow. 
The machine consists of two consecutive point clouds in time (black and red crosses), which are not allowed to move, flow arrows attached to blue points by swivel telescopic joints, and two losses (rigid and distance) represented by springs. The scene depicts two rigid vertical objects that are moving in different directions. We employ standard \emph{distance loss} (red springs), which attracts the flow of black points toward the corresponding red points from the consecutive point cloud. Since objects in the black point cloud are close to each other, no spatial clustering would separate them correctly. In contrast to existing methods, we instead introduce overlapping clusters (blue/green/magenta regions), within which the rigidity is encouraged through \emph{rigidity loss} (blue/green/magenta springs). This rigid loss constructs springs among the arrow end-points in each source rigid-clusters, which prevents the flow from deforming the rigid cluster. The color of the springs matches the color of the cluster.

We also enable outlier rejection through spectral clustering as proposed in~\cite{vidanapathirana2023mbnsf}.
In terms of the mechanical machine analogy, the intuition of outlier rejection is that it adjusts the stiffness of springs in order to weaken the springs that connect outlier points (i.e., the points, the motion of which is non-rigid with respect to the rest of the cluster).  To further regularize the flow, a structural regularization through Neural prior~\cite{li2021neural,vidanapathirana2023mbnsf,Li2023Fast,Chodosh2023reevaluating} has also been proposed. We identified that the Neural prior regularizer helps mainly with scene flow estimation in static scenes with a small number of dynamic objects; however, it over-regularizes the flow in highly dynamic scenes due to the limited capacity of the Neural-prior network. Therefore, we decided to drop this regularization in our proposed approach.


The main novelty of our approach stems from replacing the \emph{large spatially compact non-overlapping clusters} delivered by an independent clustering algorithm, such as DBSCAN~\cite{dbscan}, with two different types of clusters: (i) \emph{soft mutually overlapping clusters}, which are assumed to spread over multiple rigid objects and (ii) \emph{hard non-overlapping rigid clusters}, which are expected to cover only the points from a single rigid object or its part. The soft clusters are optimized by the rigidity loss with outlier rejection, and the hard clusters are optimized by the rigidity loss without the outlier rejection. While soft clusters remain fixed, the hard clusters are progressively growing towards consistently moving points in the spatial and temporal domain. 

We argue that while the flow estimated from large non-overlapping clusters~\cite{vidanapathirana2023mbnsf} heavily suffers from over and under-segmentation, usage of the proposed overlapping growing clusters significantly suppresses this issue, see Figure~\ref{fig:intro}. In particular, it mitigates the \emph{over-segmentation} by spatially propagating the rigidity through the overlap; see Figure~\ref{fig:losses}, for example, that the blue and magenta clusters deliver strong rigidity springs between the points of the left object, which consequently grows the hard rigid cluster over the whole rigid object. Mitigation of the \emph{under-segmentation} is achieved by using only small rigid clusters combined with outlier rejection techniques; for example, even though all three soft clusters contain an outlier, the resulting rigidity springs between left and right rigid objects are weakened, and the hard rigid cluster grows only within the rigid object. Consequently, the only assumption required in order to correctly segment inconsistently moving rigid objects is that each point of the rigid object appears as an inlier in at least one soft cluster hypothesis. 

The main contributions of the paper are three-fold:
\begin{itemize}
	\item Novel clustering method, where the rigid object segmentation is initialized from spatial-temporal grouping and is jointly optimized with the flow. 
    \item The proposed optimization-based method achieves superior accuracy among state-of-the-art methods~\cite{Li2023Fast,Chodosh2023reevaluating,li2021neural,lang2023scoop} on all tested datasets (Agroverse2~\cite{Argoverse},  Waymo~\cite{Sun_2020_CVPR}, StereoKITTI~\cite{kittisf}) 
    \item Significant improvements on long tail classes, such as independently moving pedestrians and cyclists, where current solutions fail due to structural prior regularization.
\end{itemize}

\section{Related Work}
\label{sec:related}


\paragraph{\textbf{Supervised scene flow}}
In the early stages of learning-based 3D scene flow estimation, methodologies relied on synthetic datasets~\cite{HPLFlowNet,MIFDB16,Wang_2018_CVPR, liu2019meteornet, Wang_2023_ICCV} for preliminary training.
FlowNet3D~\cite{liu:2019:flownet3d}, assimilated principles from FlowNet~\cite{FlowNet} and underwent comprehensive supervised training, employing L2 loss in conjunction with ground-truth flow annotations. 
BiPFN~\cite{BiPFN} orchestrated the bidirectional propagation of features from each point cloud, thereby enriching the representation of individual points. 
FLOT~\cite{puy20flot} introduced an innovative correspondence-based network, transforming the initial flow estimation to optimal transport estimation problem, succeeded by meticulous flow refinement through trained convolutions.
%
%


\paragraph{\textbf{Self-supervised scene flow}} Recent methodologies have sidestepped the requirement for ground-truth flow by embracing self-supervised learning. Among the early adopters of this approach, PointPWC-Net~\cite{wu2020pointpwc} presented a wholly self-supervised methodology, combining nearest-neighbors and Laplacian losses. Similarly, in \cite{Mittal_2020_CVPR}, a self-supervised nearest neighbor loss was implemented, ensuring the cycle consistency between forward and reverse scene flow.
%
RigidFlow~\cite{li2022rigidflow} formulates a methodology for generating pseudo scene flow within the domain of self-supervised learning. This approach relies on piecewise rigid motion estimation applied to collections of pseudo-rigid regions, identified through the supervoxels method~\cite{Lin2018Supervoxel}. 
%
The method known as SCOOP~\cite{lang2023scoop} adopts a hybrid framework of correspondence-based feature matching coupled with a flow refinement layer employing self-supervised losses. 
%
Compared to supervised and self-supervised approaches, our method does not require any labeled or unlabeled training data as it is optimization-based.

\paragraph{\textbf{Direct flow optimization and clustering}}
%
%
The Neural Prior\cite{li2021neural} illustrates that flows can be optimized by incorporating a structure-based prior within the network architecture.
The central objective in \cite{Li2023Fast} revolves around expediting the Neural Scene Flow~\cite{Li2023Fast}. Li \emph{et al.} identify the Chamfer distance as a computational bottleneck and use the Distance Transform as a surrogate correspondence-free loss function.
The methodology outlined in Scene Flow via Distillation by \cite{Vedder2023zeroflow} embodies a direct distillation framework.
The two newest optimization-based methods, the MBNSFP~\cite{vidanapathirana2023mbnsf} and Chodosh~\cite{Chodosh2023reevaluating} adopt Neural Prior architecture and enhanced it with spatial consistency regularization~\cite{vidanapathirana2023mbnsf} and post-process cluster rigidity~\cite{Chodosh2023reevaluating}, respectively.
Aforementioned methods use DBSCAN~\cite{dbscan} to perform clustering in single frame, whereas we use Euclidean clustering on temporally synchronized frames with bottom-up approach of connecting the points based on flows. Other related bottom-up (Agglomerative~\cite{Zepeda-Mendoza2013}) clustering approaches includes WardLinkage~\cite{Ward1963HierarchicalGT}, HDBSCAN~\cite{hdbscan}, BIRCH~\cite{birch}, which are based on mutual reachability and cluster stability. All methods use either hard clustering or hierarchical structures as a final output. They do not use overlapping clusters and flow guidance like our method.

\paragraph{\textbf{Rigidity Regularization}}
Given the presence of multiple local minima in the conventional self-supervised Chamfer distance loss, introducing regularization mechanisms becomes imperative to attain physically plausible flows—ones that faithfully trace the motion of rigid structures and objects. Implicit rigidity regularization is enforced through a robust model prior in~\cite{li2021neural,pointflownet,Yi}.
%
The introduction of weak supervision, in the form of ego-motion and foreground segmentation, has demonstrated its capacity to provide an object-level abstraction for estimating rigid flows~\cite{gojcic2021weakly3dsf}. 
%
%
%
In~\cite{Chen2022Second}, a novel global similarity measure takes the form of a second-order spatial compatibility measure on consensus seeds that serve as input to a weighted SVD algorithm, ultimately producing global rigid transformation. The mechanism was then extended to multi-object flow estimation in MBNSFP~\cite{vidanapathirana2023mbnsf}.
ICP-flow~\cite{lin2024icp} performs hierarchical clustering with histogram initialization for ICP registration per object to mitigate registration local minima. Such clustering usually suffers from the under-segmentation of objects. We differ from~\cite{lin2024icp} by utilizing per-point soft registration and specifically addressing the under-segmentation by including the clustering in the optimization process. We also differ in performing the clustering in spatial-temporal domain, rather than geometrical space, and we do not require matching the clusters in two consecutive frames.
We also use rigidity regularization, but we do not limit our method to fixed initial clusters as \cite{vidanapathirana2023mbnsf,Chodosh2023reevaluating}. Instead, we jointly optimize the initial cluster (constructed with Euclidean clustering of points in spatial-temporal space) and flow to merge and enlarge clusters while enforcing rigidity on them progressively. Another difference between our method and the aforementioned methods is the absence of Neural Prior as structural regularization, which allows us to capture correct flows of multiple mutually-exclusive object motions. 

\section{Method}
\label{sec:method}
%

The main goal is the optimization-based estimation of the flow field $F$ between two consecutive point clouds $P$ and $Q$. Point clouds consist of 3D points $\mathbf{p}_i\in P\subset\mathbb{R}^3$ and $\mathbf{q}_i\in Q\subset\mathbb{R}^3$. Since real-world scenes consist mostly of a static background, we follow a good practice of compensating the ego-motion~\cite{Chodosh2023reevaluating} first before running our method. We calculate the ICP~\cite{vizzo2023ral} as an estimate of ego motion and use it to transform the source point cloud $P$. In the rest of this section, we focus on the estimation of the remaining flow, which corresponds to the motion between transformed source point cloud $P$ and target point cloud $Q$. Consequently, the flow of point $\mathbf{p}_i \in P$ is a 3D vector $\mathbf{f}_i\in F\subset\mathbb{R}^3$ describing the velocity.

\subsection{Method overview}


The method is summarized by pseudo-code in Algorithm~\ref{alg:algorithm}. The proposed method first initializes two sets of clusters: hard and soft. \emph{Hard} rigid clusters are non-overlapping small clusters that are assumed to cover only a single rigid object or its part. Each rigid cluster $H\in\mathcal{H}$ is a small compact cluster delivered through spatial-temporal segmentation on $P$, including the adjacent temporal point clouds. In contrast, the \emph{soft} rigid clusters are small overlapping clusters that are expected to overflow into neighboring objects. Each point $\mathbf{p}\in P$ is associated with the one soft cluster $S\in \mathcal{S}$, which is defined as the set of its $k$ nearest neighbors $N_k(\mathbf{p}, P)$ from the point cloud $P$. Given these two sets of clusters, we optimize the flow to minimize (i) hard rigidity loss on hard clusters, (ii) soft rigidity loss on soft clusters, and (iii) distance loss on all points. Both rigidity losses enforce rigid flow on points within the cluster. The main difference is that the soft rigidity loss allows for outlier rejection. All losses are detailed in the following paragraphs. Once the optimized flow is available, we merge rigid clusters in $P$, whose flow goes into the same rigid cluster in $Q$. Since the algorithm is iteratively called on all consecutive pairs of point clouds, the rigidity is propagated throughout the temporal domain. The flow is initialized as a zero vector, and minimization is applied as a gradient optimization to minimize the final loss function with the Adam optimizer. The
reason for initialization in zero is that it is likely that the
nearest neighbor is going to be the same object next time
(after ICP ego-motion compensation).

\begin{algorithm*}[h!]
\caption{Joint flow estimation and rigid object segmentation}\label{alg:cap}
\begin{algorithmic}
\Require point cloud $P, Q$
\Ensure flow $F$
\begin{enumerate}
\item Initialize set of \emph{hard} rigid clusters $\mathcal{H}$. 
\item Initialize set of \emph{soft} rigid clusters $\mathcal{S}$. 
\item Minimize 
$$
\mathcal{L} = 
\alpha~
\sum_{\mathbf{f}\in F}
\mathcal{L}_{\textrm{dist}}
(\mathbf{f}) 
+ \beta\sum_{H\in \mathcal{H}}\sum_{\mathbf{f}\in H}\mathcal{L}_{\textrm{hard}}
(\mathbf{f}) 
+ \gamma \sum_{S\in \mathcal{S}}\sum_{\mathbf{f}\in S}\mathcal{L}_{\textrm{soft}}
(\mathbf{f}) 
,
$$
where $\alpha,\beta$ and $\gamma$ are hyper-parameters of the proposed method and $\mathbf{p}$, $\mathbf{q}$ and $\mathbf{f}$ are input points and corresponding flows. For simplicity, we omit per-point $\mathbf{p}$ and $\mathbf{q}$ notation.

\item If the flow of two different rigid clusters $H_i, H_j\in \mathcal{H}$ goes into the same rigid cluster in pointcloud $Q$, then merge $H_i, H_j$ into the same cluster.
\item Repeat from 3 until convergence or reaching a maximum number of iterations.
\end{enumerate}
\end{algorithmic}
\label{alg:algorithm}
\end{algorithm*}

\subsection{Distance loss}
Similarly to existing approaches, we assume that the motion of objects is sufficiently small with respect to the spatio-temporal resolution of the sensor. This assumption transforms into so-called \emph{distance loss}, which attracts the flow of points from the point cloud $P$ toward the consecutive point cloud $Q$. 
\begin{equation}
    \mathcal{L}_{\mathrm{dist}}= \sum_{i\in P} \| \mathbf{p}_i+\mathbf{f}_i - N_1(\mathbf{p}_i + \mathbf{f}_i, Q)\| 
\end{equation}
where $N_1(\mathbf{p}, Q)$ is the  nearest neighbour of point $\mathbf{p}$ from pointcloud $Q$. In practice, we use it bidirectionally for both point sets $P, Q$, which is equivalent to Chamfer distance.



\subsection{Hard rigidity loss}
Since the distance loss is typically insufficient for a reliable flow estimation, the additional prior assumption that takes into account the rigidity of objects is considered. In contrast to existing approaches~\cite{dbscan,gojcic2021weakly3dsf,song2022ogc,vidanapathirana2023mbnsf}, we do not explicitly model a fixed number of independently moving rigid objects, but we simultaneously optimize flow with the rigid clusters as describe in Algorithm~\ref{alg:algorithm}. Given a cluster $C$ consisting of $k$ points from point cloud $P$, we construct the complete undirected graph $\mathcal{G}=(\mathcal{V}, \mathcal{E})$ with $k$ vertices $\mathcal{V}=C$ and $k(k-1)/2$ edges $\mathcal{E}$ corresponding to all possible pair-wise connections among points (without self-loops). Each edge is associated with a reward function that encourages the flow in incident vertices to preserve the mutual distance between the corresponding points, \emph{i.e.}, encourage the rigid motion. The reward for preserving rigidity is defined as follows:
\begin{equation}
r_{ij} = 1- \sum_u\frac{(d_{ij}^u-\hat{d}_{ij}^u)^2}{\theta},     
\end{equation}
where $d_{ij}^u=|\mathbf{p}_{i}^u-\mathbf{p}_{j}^u|$ is the distance between the $u$-th dimension of points $\mathbf{p}_i,\mathbf{p}_j\in C$ and similarly $\hat{d}_{ij}^u=|(\mathbf{p}_{i}^u+\mathbf{f}_{i}^u)-(\mathbf{p}_{j}^u+ \mathbf{f}_{j}^u)|$ is the distance after applying the estimated flow on these points, and $\theta$ is a hyper-parameter of the proposed method. If the motion is rigid, the distance difference is zero, and the reward equals one; if the distance is non-zero the reward is proportionally smaller. Given this notation, we introduce \emph{hard rigidity loss}:
\begin{equation}
\mathcal{L}_{\textrm{hard}}(\mathbf{f}) = \sum_{(\mathbf{p}_i,\mathbf{p}_j)\in \mathcal{V}} -\log (r_{ij}).     
\end{equation}
Where we also clip the values of $r_{ij}$ between 0 and 1 to prevent potential numerical instability.

\subsection{Soft rigidity loss}
Intuitively, when the cluster $C$ is incorrect (e.g., it overflows into neighboring objects), the optimization of the flow through the hard rigid loss delivers inaccurate flow. In order to enable outlier rejection, we introduce $k$-dimensional soft clustering vector $\mathbf{v}\in \mathopen[0,1\mathclose]^{(k+1)}, \|\mathbf{v}\|=1$ that is supposed to softly selects high-reward edges. This vector models how much the points are likely to be in the rigid object that is dominant within the cluster. Given this notation, we define the soft rigidity score induced by the point $\mathbf{p}_m$ as
\begin{equation}
s_m(\mathbf{v}) = \sum_{(\mathbf{p}_i,\mathbf{p}_j)\in \mathcal{V}} r_{ij}\mathbf{v}_i\mathbf{v}_j = \mathbf{v}^\top\mathbf{A}_m\mathbf{v},\label{eq:score}
\end{equation}
where matrix $\mathbf{A}_m$ consists of elements $[\mathbf{A}_m]_{i,j}=r_{ij}$. Product $\mathbf{v}_i\mathbf{v}_j$ corresponds to the spring stiffness between point $\mathbf{p}_i$ and $\mathbf{p}_j$ in the mechanical analogy from Figure~\ref{fig:intro}.
 
Given the score matrix $\mathbf{A}_m$, the optimal soft clustering is, by Raleigh’s ratio theorem, the principal eigenvector of matrix $\mathbf{A}_m$ 
\begin{equation}
\mathrm{eig}(\mathbf{A}_m) = \mathop{\arg\max}_{ \|\mathbf{v}\|=1} \mathbf{v}^\top\mathbf{A}_m\mathbf{v}.    
\end{equation}

The score~(\ref{eq:score}) for the optimal soft clustering (which is equal to the principal eigenvalue of $\mathbf{A}_m$) describes how much the flow is consistent with the rigid motion. In order to make the estimated flow more rigid under the optimal soft clustering, we introduce the \emph{soft rigidity loss}  
\begin{equation}
\mathcal{L}_{\textrm{soft}}(\mathbf{f}) = \sum_m -\log\Big(s_m\big(\mathrm{eig}(\mathbf{A}_m(\mathbf{f}))\big)\Big),     
\end{equation}
which pushes the principal eigenvalue up and consequently makes the flow on spatially compact clusters more rigid. 
The final soft rigidity is weighted by the eigenvalues of $\mathbf{A}_m$, that are aimed to decrease the effect of loss on outlier points.
%



\ifx\false
$$
\mathop{\arg\max}_{ \|\mathbf{v}\|=1} \mathbf{v}^\top\mathbf{A}_m\mathbf{v}.
$$
This problem has the following analytical solution. Lagrangian of this constrained optimization problem
$$
L(\mathbf{v},\lambda) = \mathbf{v}^\top\mathbf{A}_m\mathbf{v} + \lambda(\|1-\mathbf{v}\|)
$$
is differentiated and equaled to zero
$$
\frac{\partial L(\mathbf{v},\lambda)}{\partial \mathbf{v}} = 2\mathbf{A}_m\mathbf{v} - 2\lambda\mathbf{v}=0.
$$
The solution of this equation
$$
\mathbf{A}_m\mathbf{v} =\lambda\mathbf{v}
$$
is any eigevector of $\mathbf{A}_m$. Substitution of these eigenvectors into the original criterion reveals that the maximum is achieved for the principal eigenvector (the one associated with the biggest eigenvalue). This solution is known as Raleigh’s ratio theorem. 
We search for the principal eigenvector $\mathrm{eig}(\mathbf{A}_m)$ through the spectral decomposition.

\fi
\section{Experiments}




\paragraph{\textbf{Datasets}}
We conducted experiments using standard scene flow benchmarks. Firstly, we employed two large-scale, well-known LiDAR autonomous driving datasets: Argoverse1~\cite{Argoverse} and the newer version Argoverse2~\cite{Wilson2021Argoverse2N}
and Waymo~\cite{Sun_2020_CVPR} open dataset. These datasets encompass challenging dynamic scenes with various maneuvers captured by different LiDAR sensor suites.
To ensure a fair comparison, we sampled and processed the LiDAR datasets according to the methodology outlined in~\cite{li2021neural} and used the first frames of all available validation sequences as done in~\cite{li2021neural} for benchmarking. 
As there are no official scene flow annotations, we adopted the data processing approach from \cite{ScalableWaymo2022} to derive pseudo-ground-truth scene flow information based on object detection annotations, akin to the methodology employed in \cite{li2021neural}. Finally, we removed ground points in the height of 0.3 meters or lower for Waymo dataset, following the procedure outlined in previous works \cite{liu:2019:flownet3d,li2021neural,lang2023scoop,Li2023Fast,gojcic2021weakly3dsf,vidanapathirana2023mbnsf}. For Argoverse1 and Argoverse2, we used the accompanying
ground maps to remove ground points and constrain the range of input points to 35 meters as done in~\cite{Chodosh2023reevaluating, gojcic2021weakly3dsf, wu2020pointpwc, lang2023scoop}.

Subsequently, we utilized the stereoKITTI dataset~\cite{kittisf, liu:2019:flownet3d}, which comprises real-world scenes from autonomous driving scenarios. The point clouds in the stereoKITTI dataset are, however, constructed by lifting the depth from stereo images by calculated optical flow, resulting in 3D points with one-to-one correspondence and a huge amount of dynamic points per object~\cite{Chodosh2023reevaluating}. The dataset was further divided by \cite{Mittal_2020_CVPR} into a testing part referred to as KITTI\textsubscript{t}.

\paragraph{\textbf{Implementation Details}} 

We use $k=16$ and set $\alpha$ and $\beta$ loss weights to 1. We use the learning rate of 0.004 and Adam optimizer. The $\theta$ distance threshold for outlier rejection is set to 0.03 as in \cite{Chen2022Second,vidanapathirana2023mbnsf}. We perform hard clustering by spatio-temporal Euclidean clustering over the horizon of 5-point clouds and group points together within the radius of 0.3m. The flows are optimized until a fixed number of iterations (1500). For other comparative methods, we use the same optimized parameters as reported in their original papers and official implementations. 

\paragraph{\textbf{Evaluation metrics}}
For proper evaluation of results, we need to define \emph{point error} $e_i$ and \emph{relative point error} $e_i^r$. Following the notation introduced in Section~\ref{sec:method}, we define these errors in meters as follows:
%
\[
e_i = \|\vect{f}_i - \vect{f}^{\text{gt}}_i\|_2 , 
\quad e^{\text{r}}_i = \frac{{\|\vect{f}_i - \vect{f}^{\text{gt}}_i\|}_2}{\|\vect{f}^{\text{gt}}_i\|_2},
\]
where $\vect{f}_i$ and $\vect{f}^{\text{gt}}_i$ are the predicted and ground-truth flow for point $\vect{p}_i$, respectively.
From $e_i, e^{\text{r}}$ we calculate Average Point Error in meters ($\textit{EPE}$), Strict Accuracy ($\textit{AS}$), Relaxed Accuracy ($\textit{AR}$), Angle Error ($\theta$) and Outliers ($Out.$), which are standard metrics used in the literature, e.g., \cite{li2021neural,lang2023scoop}. $\textit{AS}$ is a percentage of points that reached errors $e_i < 0.05$ or $e^{\text{r}}_i < 5\%$. The $\textit{AR}$ is the percentage of points for which the error satisfies either $e_i < 0.1$ or $e^{\text{r}}_i < 10\%$. Metric $Out.$ is the percentage of points with error either $e_i > 0.3$ or $e^{\text{r}}_i > 10\%$, and finally $\theta$ denotes the mean angle error between $\vect{f}_i$ and $\vect{f}^{\text{gt}}_i$. 

In automotive scenes, the majority of the points come from a static background. When the metrics ($\textit{EPE}$, $\textit{AS}$, $\textit{AR}$) are calculated and averaged for the full point cloud, the results show mostly performance on static points, lowering the importance of dynamic objects~\cite{Chodosh2023reevaluating, Vedder2023zeroflow}. If we dig even further, the dynamic points are heavily dominated by large vehicles, and the performance on smaller yet equally important classes, like Pedestrians and Cyclists, are not observable from the dynamic $\textit{EPE}$ metrics. Therefore, we also report the metrics per class on the main benchmark, Argoverse2~\cite{Wilson2021Argoverse2N}.

%
%
%
%
\begin{figure*}[t]
    \centering
    \includegraphics[width=0.99\linewidth]{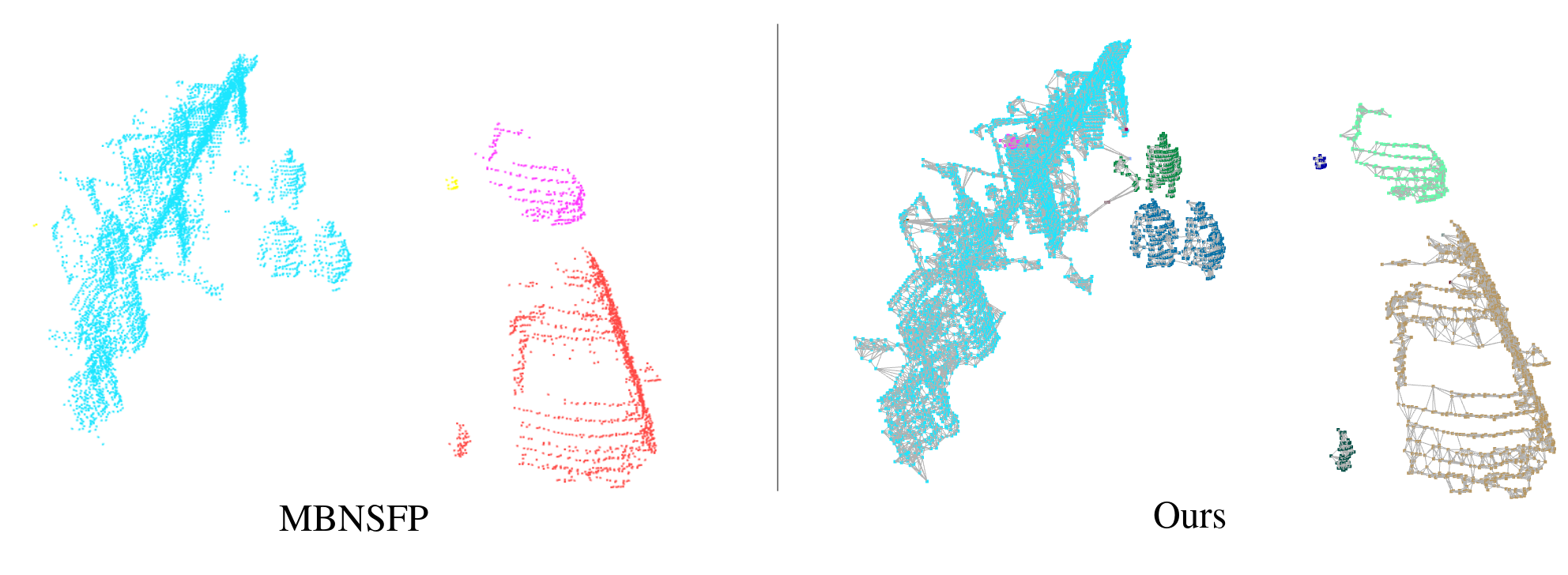}
    \caption{\textbf{On the left} - clusters from method \cite{vidanapathirana2023mbnsf}, where each color denotes a single DBSCAN cluster meant for a single rigid motion fit with outlier rejection. \textbf{On the right} - our over-segmentation Euclidean clustering (each cluster denoted by color) and soft rigidity connections that can overflow into more objects but can be rejected as outliers. We observe that a building with three pedestrians clustered can only allow for a single rigid motion (stationary from more building points) and reject the pedestrian motion in the loss for the previous method. On the other hand, motions on the right are treated separately for each object, with the exception of two pedestrians walking right next to each other.}
    \label{fig:clusters}
\end{figure*}
\subsection{Comparison with State-of-the-Art Methods}
\begin{table*}[ht]
    \centering
    \caption{Performance on Argoverse2 dataset in foreground aware metrics, i.e. Threeway EPE metric. Methods sorted by Threeway EPE Average.}
    \begin{tabular}{c|c|ccc|ccc|ccc}
    \hline
    \multicolumn{2}{c}{ } & \multicolumn{3}{|c|}{\textbf{Dynamic Foreground}} & \multicolumn{3}{|c|}{\textbf{Static Foreground}} & \multicolumn{3}{|c|}{\textbf{Static Background}} \\ 
    \toprule
    Methods    & \textit{Avg. EPE}\,[m]$\downarrow$ &  \textit{EPE}\,[m]$\downarrow$ & \textit{AS}\,[\%]$\uparrow$ & \textit{AR}\,[\%]$\uparrow$ &  \textit{EPE}\,[m]$\downarrow$ & \textit{AS}\,[\%]$\uparrow$ & \textit{AR}\,[\%]$\uparrow$ & \textit{EPE}\,[m]$\downarrow$ & \textit{AS}\,[\%]$\uparrow$ & \textit{AR}\,[\%]$\uparrow$  \\
    \toprule
    MBNSFP~\cite{vidanapathirana2023mbnsf} & 0.159 & 0.393 & 9.325 & 25.72 & \textbf{0.034} & \textbf{88.54} & \textbf{96.98} & 0.051 & 84.96 & 92.59 \\
    NSFP~\cite{li2021neural} & 0.083 & 0.141 & 39.85 & 71.69 & 0.059 & 75.15 & 91.14 & 0.048 & 81.96 & 93.56 \\
    ICP-flow~\cite{lin2024icp} & 0.078 & 0.165 & 48.61 & 70.70 & 0.039 & 79.31 & 95.32 & 0.032 & 86.68 & 95.68
    \\
    Chodosh~\cite{Chodosh2023reevaluating} & 0.070 & 0.132 & 41.80 & 75.49 & 0.049 & 77.29 & 93.74 & 0.028 & 87.97 & 95.30 \\
    \hline
    Ours & \textbf{0.047} & \textbf{0.079} & \textbf{67.90} & \textbf{85.35} & 0.035 & 86.26 & 95.78 & \textbf{0.026} & \textbf{93.02} & \textbf{96.30} \\
    \bottomrule
    \end{tabular}
    \label{tab:main-metric}
\end{table*}
\begin{table*}[ht]
    \centering
    \caption{Performance on Argoverse2 dataset on object-class aware EPE.}
    \resizebox{1.50\columnwidth}{!}{
    \begin{tabular}{c|c|c|c|c|c|c|c|c|c|c|c}
    \hline
     \multicolumn{2}{c}{} &    \multicolumn{3}{|c|}{\textbf{Pedestrian}} & \multicolumn{3}{|c|}{\textbf{Cyclist}} & \multicolumn{3}{|c|}{\textbf{Vehicles}} \\ 
    \toprule
    Methods & Neural Prior & \textit{Avg.}$\downarrow$ & \textit{Dyn.}$\downarrow$ & \textit{Stat.}$\downarrow$ & \textit{Avg.}$\downarrow$ & \textit{Dyn.}$\downarrow$ & \textit{Stat.}$\downarrow$ & \textit{Avg.}$\downarrow$ & \textit{Dyn.}$\downarrow$ & \textit{Stat.}$\downarrow$ \\
    \toprule
    MBNSFP~\cite{vidanapathirana2023mbnsf} & \cmark & 0.071 & 0.115 & 0.026 & 0.302 & 0.570 & 0.034 & 0.245 & 0.457 & \textbf{0.034} \\
    NSFP~\cite{li2021neural} & \cmark & 0.062 & 0.080 & 0.044 & 0.058 & 0.099 & 0.017 & 0.111 & 0.156 & 0.065 \\
    Chodosh~\cite{Chodosh2023reevaluating} & \cmark & 0.068 & 0.083 & 0.052 & 0.047 & 0.083 & 0.011 & 0.100 & 0.145 & 0.054 \\
    \bottomrule
    Ours & \xmark & \textbf{0.031} & \textbf{0.039} & \textbf{0.023} & \textbf{0.012} & \textbf{0.016} & \textbf{0.009} & \textbf{0.068} & \textbf{0.097} & 0.039 \\
    \bottomrule
    \end{tabular}
    }
    \label{tab:class-metric}
\end{table*}
We benchmark our method against the top-performing methods, such as NSFP~\cite{li2021neural}, MBNSFP~\cite{vidanapathirana2023mbnsf} and Chodosh~\cite{3D-MOD}, which all share the same structural optimization-based regularization, i.e Neural Prior~\cite{li2021neural,Li2023Fast}. For all methods, we use the official implementation provided in the papers. For Chodosh~\cite{Chodosh2023reevaluating}, we used the codebase published in \cite{Vedder2023zeroflow}. For each method, we first compensate the ego-motion with KISS-ICP~\cite{vizzo2023ral}, then estimate the flow by methods.

In Table~\ref{tab:main-metric}, we show results on Argoverse2, the main benchmark for 3D scene flow estimation. Our method dominates the dynamic foreground metrics, while the MBNSFP~\cite{vidanapathirana2023mbnsf} has a better static foreground. We suggest that their reliance on Neural prior architecture coupled with under-segmented rigidity tends to overfit on the static classes since they share a single movement caused by ego-motion, and the MLP layers in Neural Prior do not have the expressive capacity to catch multiple motion patterns. Chodosh~\cite{Chodosh2023reevaluating} does not optimize rigidity and Neural prior jointly but fine-tunes the rigidity as a post-processing step, resulting in more separable object motion. Our method without Neural Prior and merging clusters achieves the best average EPE over the dynamic and static metrics.

When zooming in on per-class scene flows, we observe a trend of overfitting to larger dynamic objects, i.e., vehicles. We suggest that fitting one structural neural prior to multiple objects in the scene leads to the local minima of solving the objects with the most points and sacrificing the smaller ones. In Table~\ref{tab:class-metric}, we show the performance on Pedestrians, Cyclists and Vehicles separately. For example, we can see that the performance of the other methods on the Pedestrian class is halved, compared to ours, where the over-segmentation of the scene usually safely clusters the whole Pedestrian as one hard cluster and estimates rigid flow without structural neural constraints. We see that in the Cyclist category, our performance is even higher. We do not lose the ability to estimate flows on Vehicles, as we can separate them in our merging clustering as well. We have worse static vehicles compared to MBNSFP\cite{vidanapathirana2023mbnsf}, which we explain by overfitting into static flow.

\subsection{Neural Prior with Hard and Soft Clusters}

To compare our proposed rigidity via hard and soft clusters to the most similar methods (MBNSFP~\cite{vidanapathirana2023mbnsf} as it also uses spatial consistency with outlier rejection), we change the direct optimization of flows for neural prior architecture as in other methods~\cite{li2021neural,vidanapathirana2023mbnsf, Chodosh2023reevaluating}. We also perform experiments under the metrics proposed in MBNSFP~\cite{vidanapathirana2023mbnsf}, i.e., overall endpoint error without foreground/background split.

In Table~\ref{tab:waymo-argo}, we see the results on Waymo~\cite{Sun_2020_CVPR} and Argoverse1~\cite{Argoverse} datasets on splits used in ~\cite{vidanapathirana2023mbnsf}. We can see that better performance is achieved with our rigidity regularization. The MBNSFP~\cite{vidanapathirana2023mbnsf} is the second top-performing method with regular Neural prior~\cite{li2021neural} with cycle consistency and structural regularization behind.

%
\begin{table}[t]
    \centering
    \caption{Performance on LiDAR datasets. We show performance on standard LiDAR benchmarks with overall metrics (without dynamic/class split). Results are averages over 3 runs.}
    \resizebox{\columnwidth}{!}{
    \begin{tabular}{c|c|lccc}
    \toprule
         Dataset & Method & \textit{EPE}\,[m]$\downarrow$ & \textit{AS}\,[\%]$\uparrow$ & \textit{AR}\,[\%]$\uparrow$ & $\theta$\,[rad]$\downarrow$ \\
    \midrule
    & JGF~\cite{Mittal_2020_CVPR}  & 0.542 & 8.80 & 20.28 & 0.715 \\
    & PointPWC-Net~\cite{wu2020pointpwc}  & 0.409 & 9.79 & 29.31 & 0.643 \\
    Argoverse1  & NSFP &  0.065 & 77.89 & 90.68 & \textbf{0.230} \\
    & MBNSFP~\cite{vidanapathirana2023mbnsf}  & 0.057 &  86.76 &  92.46 &  0.273 \\
    & NSFP + \textbf{Ours} & \textbf{0.050} &  \textbf{87.06} &  \textbf{94.26} & 0.269 \\
    \midrule
    %
    \midrule
    & R3DSF\cite{gojcic2021weakly3dsf}  & 0.414 & 35.47 & 44.96 & 0.527 \\
     & NSFP\cite{li2021neural}  & 0.087 & 78.96 & 89.96 & 0.300\\
    Waymo & MBNSFP\cite{vidanapathirana2023mbnsf}  & 0.066 & 82.29 & 92.44 & \textbf{0.277} \\
     & NSFP + \textbf{Ours} & \textbf{0.039} & \textbf{88.96} & \textbf{95.65} & 0.297\\
    \bottomrule
    \end{tabular}
    }
    \label{tab:waymo-argo}
\end{table}

\subsection{Results on StereoKITTI benchmark}
We also evaluate our method on both the full StereoKITTI and its testing subset \KITTIt ~ following the experimental setting in \cite{lang2023scoop,Mittal_2020_CVPR,FlowStep3D}. We compare our method to the top-performing methods in self-supervised 3D scene flow and also to recent fully-supervised methods trained on FT3D~\cite{FruhwirthReisinger2021FAST3DFS} dataset.
Our self-supervised method without any training data is on par with the most recent and top performing fully-supervised method, the IHNet~\cite{Wang_2023_ICCV} in terms of $AS$ (Ours 98.0\% and fully-supervised IHNet 97.8\%).

Our main baselines for performance comparison are state-of-the-art self-supervised methods such as SCOOP~\cite{lang2023scoop}, SLIM~\cite{Baur2021ICCV}, FastNSF~\cite{Li2023Fast}, Rigid Flow~\cite{li2022rigidflow}. Our method achieves the new self-supervised state-of-the-art performance on stereoKITTI benchmark with the relative improvement of 38\% of $EPE$ compared to top-performing SCOOP~\cite{lang2023scoop} on StereoKITTI and also achieves the improvement of 42\% of $EPE$ compared to the top-performing NSFP~\cite{Li2023Fast} on \KITTIt.
\begin{table}[t]
  \centering
  \caption{Comparison with self-supervised State-of-the-Art methods on StereoKITTI dataset. We evaluate scene flow based on standard metrics $\textit{EPE}$, $\textit{AS}$, $\textit{AR}$, and $\textit{Out.}$ used in StereoKITTI benchmarking. The previous state-of-the-art performance is underlined, and the current best is in bold. Our results averaged over five runs.}
    \resizebox{0.99\columnwidth}{!}{
    \begin{tabular}{c|c|lcccc}
    \toprule
    Test data & Method &  $\textit{EPE}(m)$ $\downarrow$ & $\textit{AS}(\%)$ $\uparrow$ & $\textit{AR}(\%)$ $\uparrow$ & $\textit{Out.}(\%)$ $\downarrow$ \\
    \midrule
    \multirow{6}*{\KITTIo} 
    & FlowStep3D~\cite{FlowStep3D} &  0.102 & 70.8 & 83.9 & 24.6 \\
    & RigidFlow~\cite{li2022rigidflow}  & 0.102 & 48.4 & 75.6 & 44.2 \\
    & SLIM~\cite{Baur2021ICCV}  & 0.067 & 77.0 & 93.4  & 24.9 \\
    & MBNSFP~\cite{vidanapathirana2023mbnsf}  & 0.112 & 80.7 & 86.3 & 14.5  \\
    & NSFP~\cite{Li2023Fast}   & 0.051 & 89.8 &  95.1 & 14.3 \\
    & SCOOP~\cite{lang2023scoop}  & \underline{0.047} & \underline{91.3} & \underline{95.0} & \underline{18.6} \\
    & \textbf{Ours}  &   \textbf{0.029} & \textbf{98.0} &  \textbf{98.7} &  \textbf{11.7} \\
      
    \midrule
    \midrule
    \multirow{6}*{\KITTIt}
    & JGF~\cite{Mittal_2020_CVPR}  &  0.105 & 46.5 & 79.4 & - \\ 
    & RigidFlow~\cite{li2022rigidflow} & 0.117 & 38.8 & 69.7 & - \\
    & MBNSFP~\cite{vidanapathirana2023mbnsf}   &  0.112 & 81.6 & 87.3 & 15.0  \\
    & SCOOP~\cite{lang2023scoop}  &   0.039 & \underline{93.6} & \underline{96.5} & 15.2 \\
    & NSFP~\cite{Li2023Fast}   & \underline{0.036} & 92.3 & 96.2 & \underline{13.2} \\%
    & \textbf{Ours} &  \textbf{0.021} &  \textbf{98.7} &  \textbf{99.1} &  \textbf{11.0} \\
    
    \bottomrule
    \end{tabular}
    }
  \label{tab:stereo}
\end{table}

\subsection{DBSCAN Clustering in Scene Flow}

In order to achieve the best possible clustering in competing methods, the authors of~\cite{vidanapathirana2023mbnsf, Chodosh2023reevaluating} tune The DBSCAN algorithm to catch the bigger objects, which is visible in parameter epsilon (distance grouping threshold) set to 0.8 and minimal number of point samples in one density-based cluster set to 30 in MBNSFP~\cite{vidanapathirana2023mbnsf}. Such parametrization allows for the successful grouping of points in whole bodies of large objects such as vehicles but simultaneously groups multiple smaller objects together. Another issue stems from points that fall below such threshold and, therefore, do not form the cluster and are treated without rigidity regularization in the methods. See an example of such DBSCAN clustering in Figure~\ref{fig:clusters}, where the multiple pedestrians are grouped together with a building. 
\begin{figure}[t]
    \centering
    \includegraphics[width=0.70\linewidth]{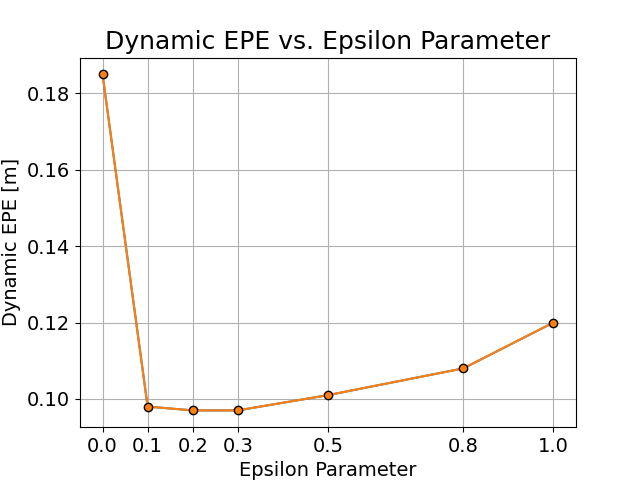}
    \caption{The hard clustering parameter Epsilon which denotes the grouping radius for Euclidean clustering compared to dynamic end-point-error.}
    \label{fig:eps-plot}
\end{figure}
We conclude that such clustering parametrization leads to over-fitting on large objects. Therefore we designed our hard clustering approach to assign all points to clusters in a "over-segmentation" manner and gradually merge the over-segmented clusters based on flow optimization and soft-clusters, see Figure~\ref{fig:eps-plot} for effect of grouping radius on dynamic object flow performance.

\subsection{Ablation Study}
In Table~\ref{tab:components}, we show metrics when adding one enhancement per time. We observe the benefits of components on the final performance on the Argoverse2 dataset.
Without rigidity regularization and clusters \textbf{a)}, the method resembles flow only to the nearest neighbors in the target point cloud and, therefore, fails to produce meaningful motions. By including the hard cluster rigidity \textbf{b)}, we observe physically consistent motions. However, since we enforce rigidity on over-segmented clusters, the objects still deform. With the addition of soft clusters \textbf{c)}, we are able to connect nearby points from different hard segments to rigidity terms and reject outliers, expanding the object rigidity. By guiding the hard clusterization with flows \textbf{d)}, we acquire a boost by connecting lonely clusters (divided by occlusions or separated by LiDAR sampling rate) back to the major rigid body. Even though the absolute difference value is small for \textit{d)}, it is non-negligible (2.5\% relative improvement on dynamic \textit{EPE}) to the remaining error.
\begin{table*}[t]
    \centering
    \caption{Ablation study for individual method components on Argoverse2.}
    \resizebox{1.99\columnwidth}{!}{
    \begin{tabular}{c|c|c|c|c}
        \toprule
        Module & Avg. $\textit{EPE}(m)$ $\downarrow$ & Dyn. Fore. $\textit{EPE}(m)$ $\downarrow$ & Stat. Fore. $\textit{EPE}(m)$ $\downarrow$ & Stat. Back. $\textit{EPE}(m)$ $\downarrow$\\
        \midrule
        a) w/o Rigidity & 0.244 & 0.504 & 0.103 & 0.124\\
        b) w/ Hard Clusters Rigidity & 0.052 & 0.089 & 0.040 & 0.028 \\
        c) w/ Hard Clusters and Soft Clusters Rigidity & 0.047 & 0.079 & \textbf{0.035} & 0.026 \\
        d) w/ Rigidity and Flow Guided Hard Clusters & \textbf{0.046} & \textbf{0.077} & \textbf{0.035} & \textbf{0.025} \\
        \bottomrule
    \end{tabular}
    \label{tab:components}
    }
    \end{table*}

\subsection{Runtime Benefit of Implementing Soft Clusters as Neighborhoods}
We analyze the performance-time trade-off in Figure~\ref{fig:time-analysis} by acquiring the $EPE$ based on inference time on Argoverse dataset while using Neural Prior regularizer with proposed Soft Clusters to demonstrate runtime difference between iterative per-cluster rigidity~\cite{vidanapathirana2023mbnsf} and neighborhood parallelization ($\mathcal{L}_{soft}$). The speed was measured on NVIDIA Tesla V100 GPU on complete point clouds. We compare with optimization-based methods FastNSF~\cite{Li2023Fast} and MBNSF~\cite{vidanapathirana2023mbnsf} with distance transform acceleration proposed by~\cite{Li2023Fast} and DBSCAN clustering, and Neural prior~\cite{li2021neural} without distance transform acceleration~\cite{Li2023Fast}. 

Compared to the MBNSFP~\cite{vidanapathirana2023mbnsf}, our method is approximately $10\times$ faster while stopping at the same prediction error. Even though their spatial consistency regularization uses the same algorithm of computing the point-to-point displacements with outlier rejection, our method is significantly faster because we design the clusters on the same-sized neighborhoods, which can be efficiently parallelized on GPUs. In contrast to our method, MBNSFP~\cite{vidanapathirana2023mbnsf} computes eigenvectors of a matrix with variable-sized clusters~\cite{dbscan}. Thus, MBNSFP~\cite{vidanapathirana2023mbnsf} is looping over $c$ clusters $\{N_0, N_1, ..., N_c\}$ one-by-one, whereas our method allows GPU to parallel the $\mathbb{R}^{N\times k \times k}$ tensor computation for $N$ points in point cloud and $k$ neighbors simultaneously. Method Without any spatial regularization, i.e. FastNP~\cite{Li2023Fast}, converges sooner with worse final performance.
\begin{tikzpicture}
    \draw (0, 0) node[inner sep=0]{
    \centering \includegraphics[width=0.99\linewidth]{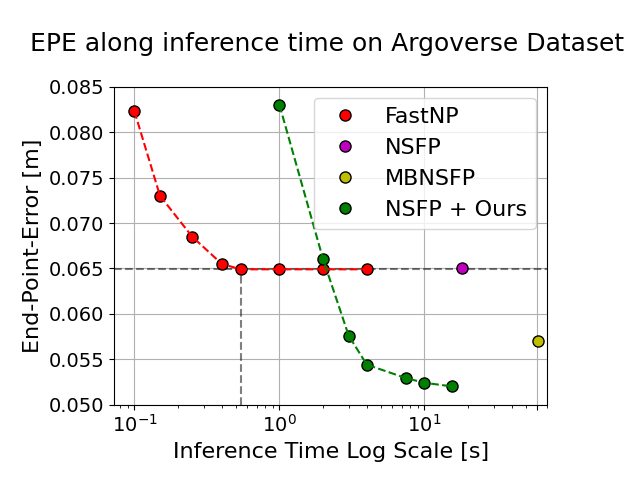}}
    ;
    \draw (2.55, 1.71) node {\cite{Li2023Fast}};
    \draw (2.55, 1.29) node {\cite{li2021neural}};
    \draw (2.55, 0.81) node {\cite{vidanapathirana2023mbnsf}};
\end{tikzpicture}
\captionof{figure}{Speed performance comparison between the optimization-based methods with Neural Prior regularizer. We show ratio between the performance and time spent of iterations. The Fast Neural Prior~\cite{Li2023Fast} converges fastest and Our proposed loss achieves the best performance. All measured with accelerator proposed in~\cite{Li2023Fast}.}
\label{fig:time-analysis}
\subsection{Limitations}
Even though our method is reasonably fast in terms of rigidity regularization, it still does not achieve real-time performance while keeping high accuracy. Potential direction for improvement is shown in~\cite{Vedder2023zeroflow}. 
Secondly, our method does not deal with the local minima of the distance loss, resulting in sub-optimal flows even when the object is correctly clustered. 
These two issues are worth of the future research to improve scene flow further.

%

\section{Conclusion}

We presented a novel self-supervised method for 3D scene flow prediction inspired by the rigid motion of multiple objects in the scene.
%
%
The method introduced overlapping cluster rigidity regularization, which achieves state-of-the-art on standard benchmarks and is tuned toward complex dynamic objects rather than static backgrounds.
The framework is part of the optimization-based models family and, therefore, does not require any training data.
%
%
%
The proposed solution adapts the rigidity mechanism to discover long tail object flows, showing that it is important, \emph{how} are the points paired into the regularization mechanism. 

\bibliographystyle{plain}
\bibliography{main}

\begin{IEEEbiography}[{\includegraphics
[width=1in,height=1.25in,clip,
keepaspectratio]{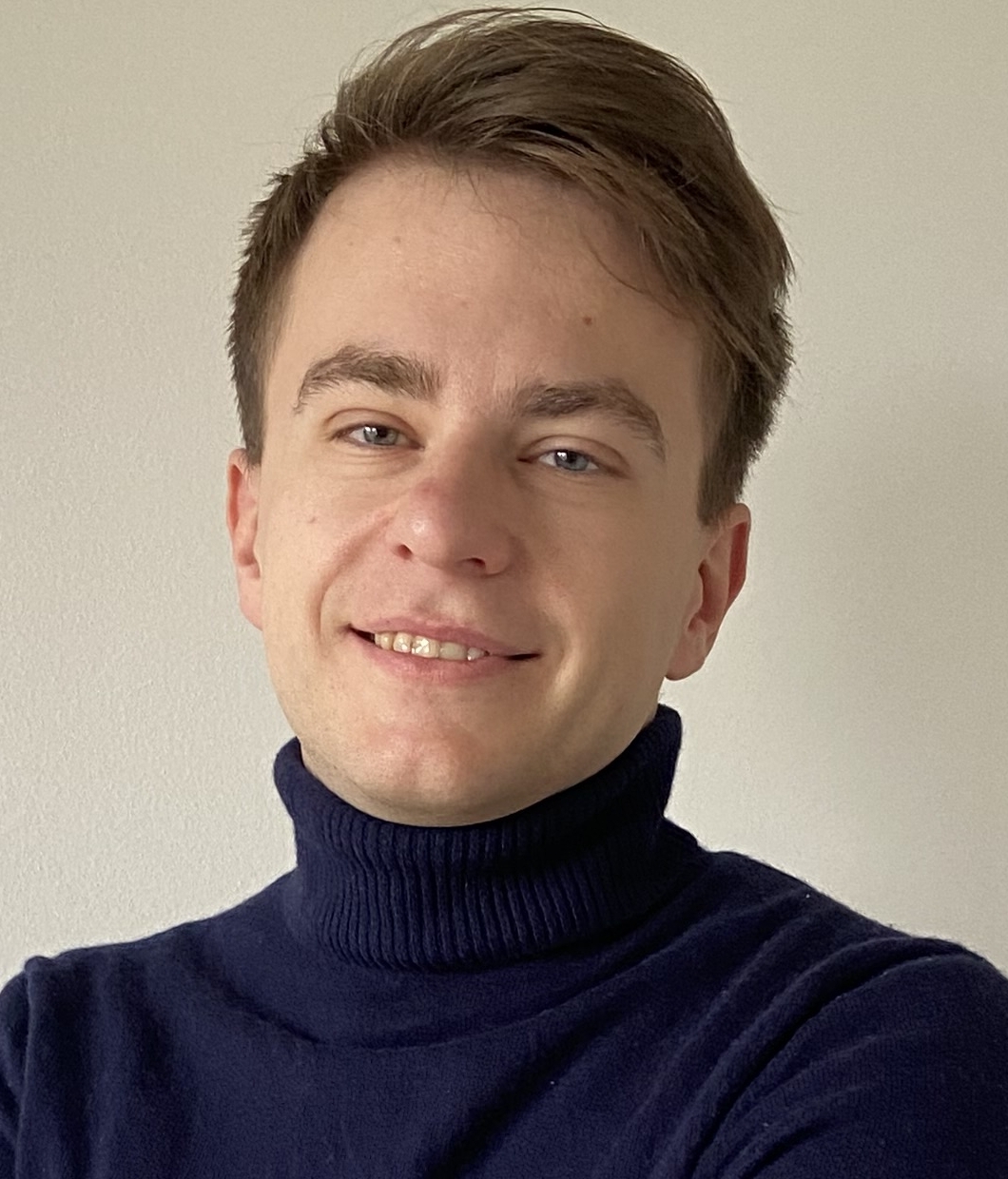}}]
{Patrik Vacek} 
received the Ing. (M.Sc.) degree in Mechatronics from Czech Technical University (CTU) in Prague in 2018. He is currently working towards the Ph.D. degree under the supervision of Tomáš Svoboda and Karel Zimmermann in the Department of Cybernetics at Faculty of Electrical Engineering of CTU, where he works as a Research Assistant. He also collaborates with Valeo R\&D centre in Prague and Paris.
\end{IEEEbiography}

\begin{IEEEbiography}[{\includegraphics
[width=1in,height=1.25in,clip,
keepaspectratio]{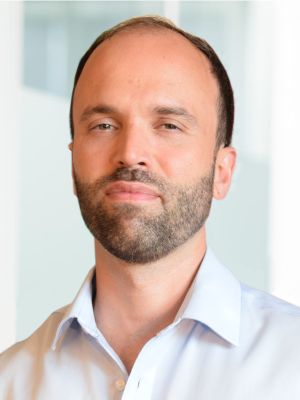}}]
{David Hurych} 
is a research scientist and senior expert in the Valeo.ai research team. His research focuses on adversarial generative models and unsupervised / semi-supervised / self-supervised machine learning methods for perception tasks on image and laser data for autonomous driving. Formerly he was a research engineer and software team leader in Valeo. David received his Ph.D. in 2014 in Biocybernetics and Artificial Intelligence study programme at the Czech Technical University in Prague.
\end{IEEEbiography}

\begin{IEEEbiography}[{\includegraphics
[width=1in,height=1.25in,clip,
keepaspectratio]{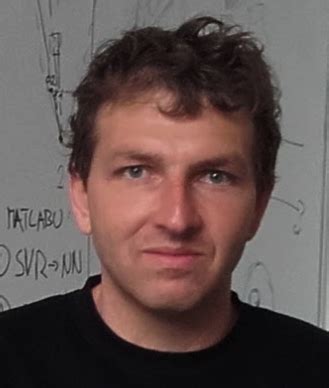}}]
{Karel Zimmermann} 
is associate professor at the Czech Technical University in Prague. He received his PhD degree in in 2008. He worked as a postdoctoral researcher with the Katholieke Universiteit Leuven (2008-2009) in the group of prof Luc van Gool. He serves as a reviewer for major journals such as TPAMI or IJCV and conferences such as CVPR, ICCV, IROS. He received the best lecturer award in 2018, the best reviewer award at CVPR 2011 and the best PhD work award in 2008. 
He was also with the Technological Education Institute of Crete (2001), with the Technical University of Delft (2002), with the University of Surrey (2006). His current research interests include learnable methods for robotics.
\end{IEEEbiography}

\begin{IEEEbiography}[{\includegraphics
[width=1in,height=1.25in,clip,
keepaspectratio]{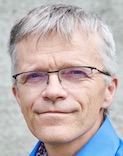}}]
{Tomáš Svoboda} is a Full Professor and Vice-dean for the development, Faculty of Electrical Engineering, Czech Technical University in Prague, the Director of the Cybernetics and Robotics PhD study program. He as published articles on multi-camera systems, omnidirectional cameras, image-based retrieval, learnable detection methods, and USAR robotics. He led the successful CTU-CRAS-NORLAB team within the DARPA SubTerranean Challenge. His research interests include multi-modal perception for autonomous systems, machine learning for better simulation and robot control, and related applications in the automotive industry and robotics.

\end{IEEEbiography}

\end{document}